\documentclass[letterpaper, 10 pt, conference]{ieeeconf}  % Comment this line out if you need a4paper

\IEEEoverridecommandlockouts                              % This command is only needed if 
\usepackage{graphicx}
\usepackage{times}
\usepackage{epsfig}
\usepackage{amsmath}
\usepackage{amssymb}
\usepackage{multirow}
\usepackage{mathtools,xparse}
\usepackage{siunitx}
\usepackage{stmaryrd}
\usepackage{amsmath}
\usepackage{hyperref}
\usepackage[noend]{algpseudocode}
\usepackage{placeins}
\usepackage{mdframed}
\usepackage{multicol}
\usepackage{amsmath,amssymb} % define this before the line numbering.
\usepackage{color}
\usepackage{subcaption}
\usepackage[table]{xcolor}

\usepackage{listings}
\usepackage{color} %red, green, blue, yellow, cyan, magenta, black, white
\definecolor{mygreen}{RGB}{28,172,0} % color values Red, Green, Blue
\definecolor{mylilas}{RGB}{170,55,241}

\algnewcommand\algorithmicforeach{\textbf{for each}}
\algdef{S}[FOR]{ForEach}[1]{\algorithmicforeach\ #1\ \algorithmicdo}

\newcommand{\trans}{\text{T}}
\newcommand{\adj}{\mathbf{\text{adj}}}
\newcommand{\matr}[1]{\mathbf{#1}}     % ISO complying version

\newcommand{\Proj}{\matr{P}}

\newcommand{\Fund}{\matr{F}}
\newcommand{\Ess}{\matr{E}}
\newcommand{\Intrinsic}{\matr{C}}

\title{\LARGE \bf
Least-squares Optimal Relative Planar Motion for\\Vehicle-mounted Cameras
}

\author{Levente Hajder$^{1}$ and Daniel Barath$^{2}$% <-this % stops a space
\thanks{*Levente Hajder was supported by the project EFOP-3.6.3-VEKOP-16-2017-00001: Talent Management in Autonomous Vehicle Control Technologies, by the Hungarian Government and co-financed by the European Social Fund. His work was also supported by Thematic Excellence Programme, Industry and Digitization Subprogramme, NRDI Office, 2019. Daniel Barath was supported by the Hungarian Scientific Research Fund (No.\ NKFIH OTKA KH-126513 and K-120499) and by the OP VVV project CZ.02.1.01/0.0/0.0/16019/000076 Research Center for Informatics.
}% <-this % stops a space
\thanks{$^{1}$Levente Hajder is with the Department of Algorithms and their Applications,
        Eotvos Lorand University, Budapest, Hungary
        {\tt\small hajder@inf.elte.hu}}%
        \thanks{$^{2}$Daniel Barath is with Visual Recognition Group, Department of Cybernetics, Czech Technical University in Prague, Czech Republic and with the
        Machine Perception Research Laboratory, MTA SZTAKI, Budapest, Hungary 
        {\tt\small barath.daniel@sztaki.mta.hu}}%
}

\begin{document}

\maketitle
\thispagestyle{empty}
\pagestyle{empty}

%%%%%%%%%%%%%%%%%%%%%%%%%%%%%%%%%%%%%%%%%%%%%%%%%%%%%%%%%%%%%%%%%%%%%%%%%%%%%%%%
\begin{abstract}
A new closed-form solver is proposed minimizing the algebraic error optimally, in the least squares sense, to estimate the relative planar motion of two calibrated cameras.
The main objective is to solve the over-determined case, i.e., when a larger-than-minimal sample of point correspondences is given -- thus, estimating the motion from at least three correspondences.
The algorithm requires the camera movement to be constrained to a plane, e.g.\ mounted to a vehicle, and the image plane to be orthogonal to the ground.\footnote{Note that the latter constraint can be straightforwardly made valid by transforming the image when having the gravity vector, e.g., from an IMU.}
The solver obtains the motion parameters as the roots of a $6$th degree polynomial.
It is validated both in synthetic experiments and on publicly available real-world datasets that using the proposed solver leads to results superior to the state-of-the-art in terms of geometric accuracy with no noticeable deterioration in the processing time. 
\end{abstract}

%%%%%%%%%%%%%%%%%%%%%%%%%%%%%%%%%%%%%%%%%%%%%%%%%%%%%%%%%%%%%%%%%%%%%%%%%%%%%%%%
\section{INTRODUCTION}

The estimation of the epipolar geometry between a stereo image pair is a fundamental problem of computer vision for recovering the relative camera motion, i.e.\ the rotation and translation of the cameras. 
Being a well-studied problem, several papers discussed its theory, estimators and potential applications. 
Faugeras proved~\cite{Faugeras1992} that this relationship is described by a $3 \times 3$ projective transformation: this is the so-called fundamental matrix. 
When the intrinsic camera parameters are known, additional geometric constraints transform it to an essential matrix. 
In this paper a special constraint is considered to hold, i.e.\ the planar motion, when the 
optical axes of the cameras are in the same 3D plane and their vertical directions are parallel.
A new solver is proposed for estimating the camera rotation and translation formalizing the problem as a least-squares optimization.
In particular, \textit{we are interested in solving optimally the over-determined case}, i.e.\ to estimate the camera motion from a larger-than-minimal set of point correspondences.

\begin{figure}[h]
  	\centering
  	\includegraphics[width=0.95\columnwidth]{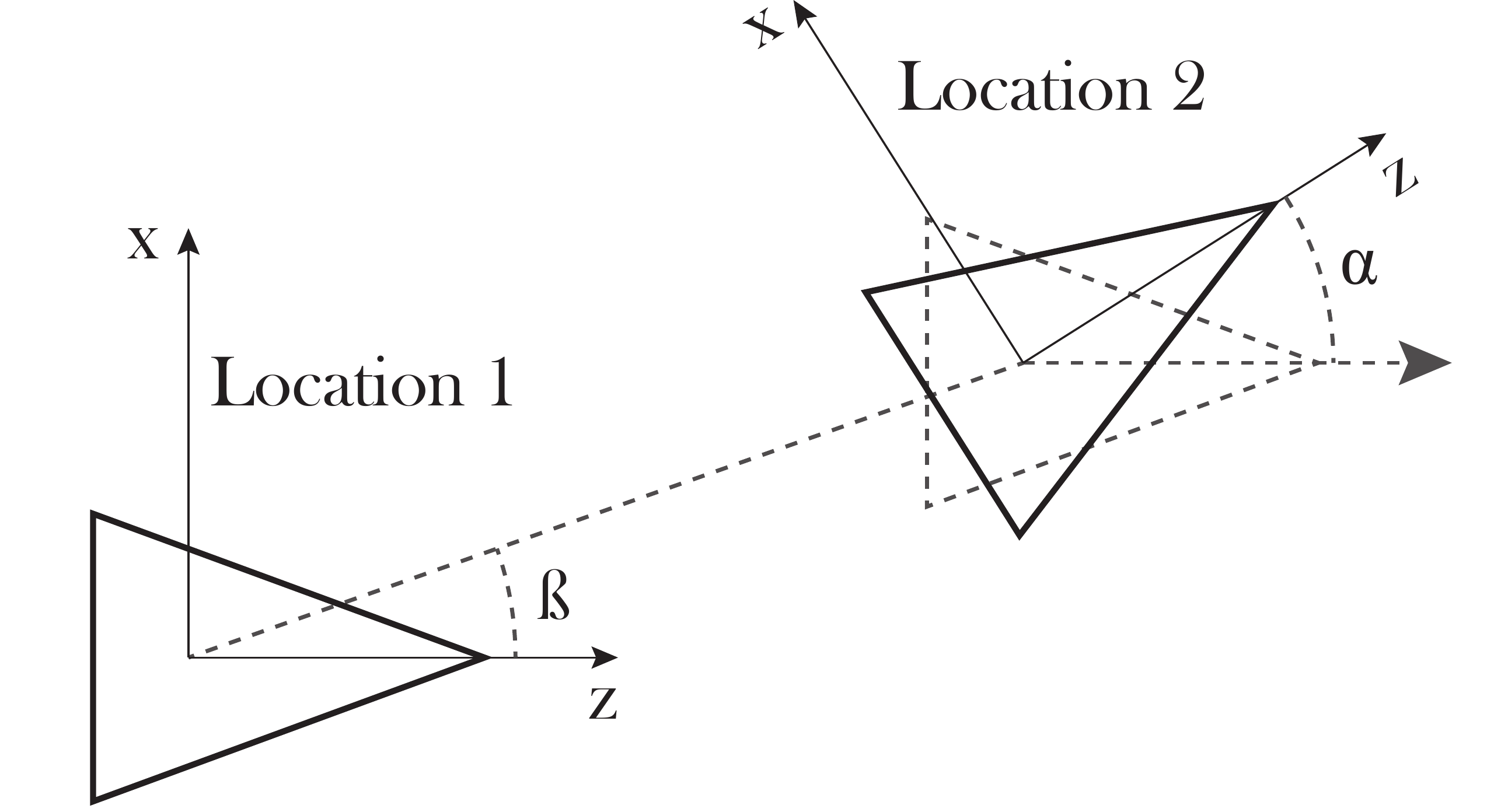}
  	\caption{\textit{Planar motion scheme.} The camera movement is described by the angle $\alpha$ of the rotation perpendicular to axis $\text{Y}$ and translation vector $[\cos(\beta), \; 0 \;, \sin(\beta)]^\trans$. }
	\label{fig:illustration_image}
\end{figure}

There is a number of techniques proposed for estimating the epipolar geometry from different numbers of correspondences considering general camera movement. For instance, such methods are the five-~\cite{philip1996non,nister2004efficient,li2006five,batra2007alternative}, six-~\cite{stewenius2006recent,li2006simple,wang1997six,hartley2012efficient}, seven- and eight-point~\cite{hartley2003multiple} algorithms. 
These methods consider no special camera movement and, therefore, are applicable in most of the situations without introducing restrictions on the cameras. They are used in most of the state-of-the-art SLAM pipelines~\cite{Cadena2016}, and fundamental methods for the motion estimation. 
They are powerful tools for estimating the pose when having a minimal set of point correspondences. 
However, the over-determined case has not yet been solved optimally while keeping the geometric constraints valid. For example, to estimate the pose from a larger-than-minimal sample by the five-point algorithm, usually, the points are fed into the null-space calculation. Then, the geometric constraints are forced to the four singular vectors corresponding to the four smallest singular values. Since those singular vectors are not null-vectors anymore, the procedure will distort the noise and result in a far-from-optimal pose estimate.
Therefore, these methods require a final numerical refinement of the pose parameters.

Nowadays, one of the most popular applications of computer vision is the sensing of autonomous vehicles. 
Algorithms designed for such problems must be robust and fast to reduce the probability of failure. 
This setting allows us to exploit the special properties of the vehicle movement to provide results superior to the state-of-the-art general algorithms both in terms of geometric accuracy and processing time. Fig.~\ref{fig:illustration_image} shows the movement of a typical vehicle-mounted monocular camera.
A straightforward way to incorporate this additional information is to put constraints on the rotation and translation parameters of the cameras, for which, several approaches exist.

\noindent
\textit{Circular planar motion.}
One of the most simplified models is the circular planar motion, i.e.\ the so-called non-holonomic constraint, which assumes that the camera undergoes planar and, also, circular motion. 
This assumption is valid if the camera is mounted to a vehicle which rotates around an exterior point defined by the intersecting lines drawn by the orientation of the wheels. 
For instance, this is the case which holds for standard cars. 
Straight driving is interpreted as rotating around a point in the infinity, i.e.\ movement on a circle with infinite radius.
Scaramuzza~\cite{Scaramuzza2011} proposed a one-point technique considering this kind of movement. Choi et al.~\cite{choi2011does} improved the accuracy by normalization.
Civera et al.~\cite{civera20101} proposed a technique for real-time visual odometry and structure-from-motion using extended Kalman-filtering. 
Interestingly, this motion assumption is a way of resolving the scale ambiguity~\cite{scaramuzza20111} if the camera is not located over the rear axle.
Even though this circular planar motion constraint holds for a number of moving vehicles in ideal circumstances, it is not able to describe several movement types, e.g.\ steering of a car or robot turning in one place. 

\noindent
\textit{Planar motion.}
In general, the cameras follow planar motion when moving on planar surfaces such as roads or floors. 
Ortin and Montiel~\cite{Ortin2001} showed that, in this case, the problem becomes the estimation of two unknown angles, see Fig.~\ref{fig:illustration_image}, and can be solved using two point correspondences. 
In \cite{Ortin2001}, the authors proposed an iterative approach.        
Formalizing the problem as the intersection calculation of two geometric entities, Chou and Wang~\cite{chou2015} and Choi et al.~\cite{choi2018} proposed closed-form solutions. 
These methods obtain the pose parameters as the intersections of two ellipses, circles, and line-circle, respectively. 
Considering this kind of general planar motion is fairly often done with additional sensors installed to the vehicle.  
For example, Troiani et al.~\cite{troiani20142} proposed a two-point RANSAC for a camera-IMU system installed to micro aerial vehicles.
He et al.~\cite{he2016automated} assumes a UAV to fly on an approximately constant altitude and estimate the camera motion by the two-point algorithm. 
Zhang et al.~\cite{zhang2015orientation} present an orientation estimation approach based on a monocular camera and IMUs.
Lee et al.~\cite{hee2013motion} devised a novel two-point algorithm for generalized cameras mounted to cars. 
Choi and Park~\cite{choi2015new} exploited planar motion estimation using RGB-D sensors.
Also, when the gravity vector is known (e.g., from an IMU), the points can be straightforwardly transformed to make the planarity constraint valid. 
However, \textit{these methods cannot cope with the optimal estimation} of motion parameters \textit{when a larger-than-minimal correspondence set is given}.

In this paper, planar motion is considered and, as the main contribution, a solver is proposed approaching the problem as a minimization of a quadratic cost incorporating the two unknown angles. 
The solver is closed-form, and the solution is obtained as the roots of a $6$th order polynomial, thus the optimal estimate has to be selected from at most $6$ solutions. It is shown both in a synthetic environment and on more than $9.000$ publicly available real-world image pairs that the method is superior to the state-of-the-art in terms of geometric accuracy.  

\section{Notation and preliminaries}
\label{sec:notations}

The basic symbols and algebraic notations are written in this section.
Scalars are denoted by regular fonts, vectors and matrices by bold ones.
The proposed solver is based on computations with univariate polynomials. 
The variable is always $\lambda$. 
A polynomial is denoted by a bold uppercase character with lower and upper indices. 
The lower one denotes the index of the polynomial, the upper one is its degree. 
For instance, $\matr Q_1^3 \left( \lambda \right)$ is a cubic polynomial:
\begin{equation*}
    \matr Q_1^3 \left( \lambda \right) = q_3 \lambda^3 +q_2 \lambda^2 + q_1 \lambda +q_0 = 
    \sum_{i=0}^3 q_i \lambda^i, 
\end{equation*}
where each $q_i$ is a coefficient of the polynomial.
The multiplication of two polynomials is denoted as
\begin{equation*}
    \matr Q_1^N \left( \lambda \right)  \matr P_1^M \left( \lambda \right) = \\
        \sum_{i=0}^\text{N} \sum_{j=0}^\text{M}  q_i p_j  \lambda^{(i+j)},
\end{equation*}
where $N, M \in \mathbb{R}$ are degrees. Normally, the resulting polynomial after the multiplication is of degree $N+M$.
The squared-norm of a polynomial is defined as the multiplication of that by itself. If the degree is $N$, then that of the squared-norm is $2N$ as follows:
\begin{equation*}
    \left| \left| \matr Q_1^N \left( \lambda \right) \right| \right|_2^2   =
    \matr Q_1^N \left( \lambda \right) \matr Q_1^N \left( \lambda \right)=\matr Q_2^{2N} \left( \lambda \right) .
\end{equation*}

\begin{figure*}
\begin{equation}
\label{eq:adj}
{
\adj(\matr M)=\left[\begin{array}{ccc}\scriptstyle \left(\lambda+\matr a_{2}^{\trans}\matr a_{2}\right)\left(\matr a_{3}^{\trans}\matr a_{3}-\lambda\right)-\left(\matr a_{2}^{\trans}\matr a_{3}\right)^{2} & \scriptstyle \matr a_{2}^{\trans}\matr a_{3}\matr a_{3}^{\trans}\matr a_{1}-\matr a_{2}^{\trans}\matr a_{1}\left(\matr a_{3}^{\trans}\matr a_{3}-\lambda\right) & \scriptstyle \matr a_{2}^{\trans}\matr a_{1}\matr a_{3}^{\trans}\matr a_{2}-\matr a_{3}^{\trans}\matr a_{1}\left(\lambda+\matr a_{2}^{\trans}\matr a_{2}\right)\\ \scriptstyle
\matr a_{1}^{\trans}\matr a_{3}\matr a_{3}^{\trans}\matr a_{2}-\matr a_{2}^{\trans}\matr a_{1}\left(\matr a_{3}^{\trans}\matr a_{3}-\lambda\right) & \scriptstyle \left(\lambda+\matr a_{1}^{\trans}\matr a_{1}\right)\left(\matr a_{3}^{\trans}\matr a_{3}-\lambda\right)-\left(\matr a_{1}^{\trans}\matr a_{3}\right)^{2} & \scriptstyle \matr a_{1}^{\trans}\matr a_{2}\matr a_{3}^{\trans}\matr a_{1}-\matr a_{3}^{\trans}\matr a_{1}\left(\lambda+\matr a_{1}^{\trans}a_{1}\right)\\ \scriptstyle
\matr a_{2}^{\trans}\matr a_{3}\matr a_{3}^{\trans}\matr a_{1}-\matr a_{2}^{\trans}\matr a_{1}\left(\matr a_{3}^{\trans}\matr a_{3}-\lambda\right) & \scriptstyle \matr a_{1}^{\trans}\matr a_{2}\matr a_{3}^{\trans}\matr a_{1}-\matr a_{3}^{\trans}\matr a_{1}\left(\lambda+\matr a_{1}^{\trans}\matr a_{1}\right) & \scriptstyle \left(\lambda+\matr a_{1}^{\trans}\matr a_{1}\right)\left(\lambda+\matr a_{2}^{\trans}\matr a_{2}\right)-\left(\matr a_{1}^{\trans}\matr a_{2}\right)^{2}
\end{array}\right]}
\end{equation}
\end{figure*}

\noindent
\textit{Fundamental and essential matrices.}
The $3 \times 3$ fundamental matrix $\matr{F}$ is a projective transformation ensuring the epipolar constraint as $\matr{p_2}^\trans \matr{F} \matr{p_1} = \matr{p_2}^\trans \Intrinsic_2^{-\trans} \matr{E} \Intrinsic_1^{-1} \matr{p_1} = 0$. The relationship of essential matrix $\matr{E}$ and $\matr{F}$ is $\matr{F} = \Intrinsic_2^{-\trans} \matr{E} \Intrinsic_1^{-1}$, where matrices $\Intrinsic_1$ and $\Intrinsic_2$ contain the intrinsic parameters of the two cameras.
In the rest of the paper, we assume points $\matr{p_1}$ and $\matr{p_2}$ to be premultiplied by  $\Intrinsic_1^{-1}$ and $\Intrinsic_2^{-1}$ simplifying the epipolar constraint to 
\begin{equation}
	\label{eq:epipolar_constraint} 
	\matr{q_2}^\trans \matr{E} \matr{q_1} = 0,
\end{equation}
where $\matr{q}_1=\Intrinsic_1^{-1} \matr{p}_1= [q_{1x}\quad q_{1y} \quad 1]^\trans  $ and $\matr{q_2} =\Intrinsic_2^{-1} \matr{p_2}= [q_{2x}\quad q_{2y} \quad 1]^\trans  $ are the normalized points. Essential matrix $\matr{E}$ can be described by the camera motion as follows: $\matr{E} = [\matr{t}]_\times \matr{R}$, where $\matr{t}$ is a 3D translation vector and $\matr{R}$ is an orthonormal rotation matrix. Operator $[.]_\times$ is the cross-product matrix. 
The $i$th element of the essential $\matr{E}$ and fundamental matrices $\matr{F}$ in row-major order is denoted as $e_i$ and $f_i$, respectively, $i \in [1,9]$ as follows:
\begin{equation*}
\Ess =\left[ \begin{array}{ccc} e_1 & e_2 & e_3 \\ e_4 & e_5 & e_6 \\ e_7 & e_8 & e_9 \end{array} \right],\;
\Fund =\left[ \begin{array}{ccc} f_1 & f_2 & f_3 \\ f_4 & f_5 & f_6 \\ f_7 & f_8 & f_9 \end{array} \right].
\end {equation*}

\noindent
\textit{Planar motion.}
Suppose that a calibrated image pair with a common $\text{XZ}$ plane is given. Directions $\text{X}$ and $\text{Y}$ are the horizontal and vertical ones, while axis $\text{Z}$ is perpendicular to the image planes. Having a common $\text{XZ}$ plane means that the vertical image directions are parallel. A trivial example for that constraint is the camera setting of usual autonomous cars: a camera is fixed to the moving car, and the $\text{XZ}$ plane of the camera is parallel to the ground plane.

Let us denote the first and the second projection matrices
by $\Proj_1$ and $\Proj_2$. Without loss of generality, the world coordinate system is fixed to the first camera. Therefore, $\Proj_1$ is represented by
$
\Proj_1 = \Intrinsic_1 \; [ \; \matr{I}_{3 \times 3} \; | \; 0 \; ],
$
where $\Intrinsic_1$ is the intrinsic camera parameters of the first camera.
The second one is written as a general pinhole camera as follows:
$
\Proj_2 = \Intrinsic_2 \; [ \; \matr{R}_{2} \; | \; \matr{t}_{2} \; ],
$
where $\Intrinsic_2 $, $\matr{R}_{2}$, and $\matr{t}_{2}$ are the intrinsic camera matrix, orientation and location of the second camera, respectively.
Assuming planar motion and a common $\text{XZ}$ plane, the rotation and translation are represented by three parameters: a 2D translation and a rotation. Formally,
\[
\begin{array}{ccc}
\matr{R}_{2} = \left[\begin{array}{ccc}
 \cos  \alpha & 0 & \sin  \alpha \\
 0 & 1 & 0 \\
 - \sin  \alpha & 0 &  \cos  \alpha \\
\end{array}\right],
&~&
\matr{t} = \left[\begin{array}{c}
x\\
0\\
z
\end{array}\right].
\end{array}
\]
Describing this motion by epipolar geometric entities, the general $3 \times 3$ essential matrix
$\Ess = \left[ \matr{t} \right]_{\times} \matr{R} $ have to be modified, where $\matr{R}$ is the relative rotation between the images. In the discussed case, $\matr{R} = \matr{R}_{2}$ and 
\[
\left[ \matr{t} \right]_{\times}=\left[\begin{array}{ccc}
0 & -z & 0\\
z & 0 & -x\\
0 & x & 0
\end{array}\right].
\]
Therefore, the essential matrix is simplified for planar motion as follows: 
\begin{eqnarray*}
    \small
	\Ess = \left[ \matr{t} \right]_{\times} \matr{R}_{2} = 
	\left[\begin{array}{ccc}
	0 & -z & 0 \\ 
	z \cos \alpha + x \sin \alpha & 0 & z \sin \alpha -x \cos \alpha\\
	0 & x & 0
	\end{array}\right] 
	\label{eq:E}
\end{eqnarray*}
Thus, $e_{1} = e_{3} = e_{5} = e_{7} = e_{9} = 0$, $e_{2} = -z$, $e_{8} = x$,
$e_{4} = z \cos \alpha - x \sin  \alpha $, and $e_{6} = -z \sin  \alpha - x  \cos  \alpha$. As it is well known in projective geometry~\cite{hartley2003multiple}, the scale of the motion cannot be retrieved, only the direction. 
Therefore the planar translation parameters $x$ and $z$ are described as the coordinates of a point on unit-circle as follows: $x = \cos \beta$ and $z = \sin \beta$, where $\beta$ is an angle. Due to this, the non-zero elements of the essential matrix are rewritten as
\begin{eqnarray*}
	&e_2 = -\sin \beta , e_8 = \cos \beta , \nonumber \\
	&e_4 = \sin \beta \cos \alpha + \cos \beta \sin  \alpha = \sin \left( \beta + \alpha \right), \nonumber \\ 
	&e_6 = \sin \beta \sin  \alpha - \cos \beta  \cos  \alpha = - \cos \left( \beta + \alpha \right). 
	\label{eq:E_planar_elements}
\end{eqnarray*}
Consequently, the motion has two degrees-of-freedom: the angles of the rotation and translation.  

\section{LEAST-SQUARES OPTIMAL SOLVER}
\label{sec:all_const}

In this section a solver, optimal in the least-squares sense, is proposed for estimating the essential matrix from a non-minimal set of point correspondences (at least three) in case of planar motion.

The problem can be written as a homogeneous linear system of equations, $\matr A \matr x = \matr 0$, where 
\begin{equation*}
    \matr x=[\cos \beta \quad \sin \beta \quad \cos \left( \alpha + \beta \right) \quad \sin \left( \alpha + \beta \right)]^\trans.
\end{equation*}
The coordinates in vector $\matr x$ are dependent. The first and last two ones are the sine and cosine functions of the same angles. Also, the solution is defined only up to a scale. 
Therefore, if parameter vector $\matr x$ is divided into two 2D sub-vectors, the lengths of these vectors have to be equal.

The constraints can be added by applying Lagrangian multipliers. The scale ambiguity and equality of the two sub-vectors requires two multipliers. 
However, the scale ambiguity can be represented by fixing one of the coordinates in vector $\matr x$. 
We fix the third coordinate as follows: $\matr x = [\gamma \quad \delta \quad \epsilon \quad 1]^\trans$. 
%This parameterization is unstable if the last coordinate, i.e.\ $\sin(\alpha+\beta)$, is close to zero. 
%For this reason, another very similar parameterization has to be applied, the third coordinate of $\matr x$ must set to one: $\matr x = [\gamma \quad \beta \delta 1 \quad \epsilon]^T$. We discuss the former case here, the other parameterization can be straightforwardly handled. %
Note that in case of $\sin(\alpha+\beta)$ being close to zero, this parameterization can be unstable. Thus $\matr x = [\gamma \quad \delta \quad 1 \quad \epsilon]^\trans$ also has to be calculated. The former one is discussed here. The latter one can be straightforwardly obtained by swapping the third and fourth columns of $\matr A$. 

Let us denote the $i$th row of matrix $\matr A$ by vector $\matr a_i$, and formulate the problem with the new parameters as the minimization of cost function $J$ which is as follows:
\begin{equation}
J = \left\Vert \gamma \matr{a_{1}} + \delta \matr{a_{2}} + \epsilon \matr{a_{3}} + \matr{a_{4}}\right\Vert _{2}^{2}.
\label{eq:cost_orig}
\end{equation}
In order to enforce the constraint of the lengths of the 2D sub-vectors (i.e., $\gamma^2 + \delta^2 = \epsilon^2 + 1$), only a single Lagrangian multiplier is introduced, and the cost function is modified as
\begin{equation*}
\widehat{J} = \left\Vert \gamma \matr{a_{1}} + \delta \matr{a_{2}} + \epsilon \matr{a_{3}} + \matr{a_{4}}\right\Vert _{2}^{2} + \lambda\left(\gamma^2 + \delta^2 - \epsilon^{2}-1\right).
\end{equation*}
The minimum is obtained by the derivatives w.r.t.\ the unknown parameters $\gamma$, $\delta$, and $\epsilon$ as
\begin{eqnarray*}
\frac{\partial \widehat{J}}{\partial \gamma} & = 2 (\gamma \matr a_{1}^{\trans} + \delta \matr a_{2}^{\trans} + \epsilon \matr a_{3}^{\trans} + \matr a_{4}^{\trans}) \matr a_{1} + 2 \lambda \gamma &= 0, \\
\frac{\partial \widehat{J}}{\partial \delta} & = 2 (\gamma \matr a_{1}^{\trans} + \delta \matr a_{2}^{\trans} + \epsilon \matr a_{3}^{\trans} + \matr a_{4}^{\trans}) \matr a_{2} + 2 \lambda \delta &= 0, \\
\frac{\partial \widehat{J}}{\partial \epsilon} & = 2 (\gamma \matr a_{1}^{\trans} + \delta \matr a_{2}^{\trans} + \epsilon \matr a_{3}^{\trans} + \matr a_{4}^{\trans}) \matr a_{3} - 2 \lambda \epsilon &= 0.
\end{eqnarray*}
They can be written in matrix form as
%\begin{eqnarray*}
%\small
% \left( \left[ \begin{array}{c} \matr a_{1}^{\trans} \\ \matr a_{2}^{\trans} \\ \matr a_{3}^{\trans}   \end{array} \right]
% \left[ \begin{array}{c} \matr a_{1}^{\trans} \\ \matr a_{2}^{\trans} \\ \matr a_{3}^{\trans}   \end{array} \right]^T
 %\left[
 %\begin{array}{ccc} \right]  \matr a_{1} & \matr a_{2} & \matr a_{3} \end{array}
 %\right]
 %+
 %\text{diag} \left( %\lambda,\lambda,-\lambda \right)
% \left[ \begin{array}{ccc} \lambda & 0 & 0 \\ 0 & \lambda & 0 \\ 0 & 0 & -\lambda  \end{array}  \right]
%\right)
%\left[\begin{array}{c}
%\gamma\\
%\delta\\
%\epsilon
%\end{array}\right]= -\left[\begin{array}{c}
%\matr a^T_{1}\\
%\matr a^T_{2}\\
%\matr a^T_{3}
%\end{array}\right] \matr a_{4}.
%\end{eqnarray*}

\begin{eqnarray*}
\small
\underbrace{\left[\begin{array}{ccc}
\lambda+\matr a_{1}^{\trans}\matr a_{1} & \matr a_{1}^{\trans}\matr a_{2} & \matr a_{1}^{\trans}\matr a_{3}\\
\matr a_{2}^{\trans}\matr a_{1} & \lambda+\matr a_{2}^{\trans}\matr a_{2} & \matr a_{2}^{\trans}\matr a_{3}\\
\matr a_{3}^{\trans}\matr a_{1} & \matr a_{3}^{\trans}\matr a_{2} & \matr a_{3}^{\trans}\matr a_{3}-\lambda
\end{array}\right]}_{\matr M \left( \lambda \right)}
\underbrace{
\left[\begin{array}{c}
\gamma\\
\delta\\
\epsilon
\end{array}\right]}_{\matr x}=
\underbrace{-\left[\begin{array}{c}
\matr a^\trans_{1}\\
\matr a^\trans_{2}\\
\matr a^\trans_{3}
\end{array}\right] \matr a_{4}}_{\matr b}.
\end{eqnarray*}
In short, it can be written as a homogeneous linear system of equations: $\matr M \left( \lambda \right) \matr x= \matr b $. The elements of vector $\matr x$ can be obtained by multiplying this system with matrix $\matr M^{-1} \left( \lambda \right)$. The inverse is given by the division of the adjoint matrix with the determinant:
\begin{equation}
\left[\begin{array}{c}
\gamma \\
\delta \\
\epsilon
\end{array}\right]=-\frac{ \adj(\matr M(\lambda))}{\det(\matr M(\lambda))}\left[\begin{array}{c}
\matr a_{1}^{\trans}\\
\matr a_{2}^{\trans}\\
\matr a_{3}^{\trans}
\end{array}\right] \matr a_{4}
\label{eq:x_sol}
\end{equation}
The adjoint matrix is written in Eq.~\ref{eq:adj}. Its elements are polynomials in $\lambda$, their degrees are $1$ or $2$. 
The determinant is a cubic polynomial. 
Thus, all the elements can be expressed as the fraction of a quadratic and a cubic polynomial as 
\begin{eqnarray*}
\small
\begin{array}{ccc}
\gamma = {\matr P_{1}^{2}(\lambda)} / {\matr P_{4}^{3}(\lambda)},&
\delta = {\matr P_{2}^{2}(\lambda)} / {\matr P_{4}^{3}(\lambda)},&
\epsilon = {\matr P_{3}^{2}(\lambda)} / {\matr P_{4}^{3}(\lambda)}
\end{array} \end{eqnarray*}
Polynomials $\matr P_{1}^{2}$, $\matr P_{2}^{2}$, and $\matr P_{3}^{2}$ are obtained by multiplying the first, second, and third row of the adjoint matrix $\adj(\matr M(\lambda))$ with the vector $\left[\begin{array}{ccc}
\matr a_{4}^{\trans}\matr a_{1}&
\matr a_{4}^{\trans}\matr a_{2}&
\matr a_{4}^{\trans}\matr a_{3}
\end{array}\right]^\trans$, and $\matr P_{4}^{3}$ is the determinant of matrix $\matr M(\lambda)$.

In order to determine the Lagrange multiplier, condition $\gamma^{2} + \delta^{2} - \epsilon^{2} - 1=0$ has to be considered. 
This leads to a sixth-degree polynomial in $\lambda$ as follows: 
\begin{equation*}
\left( \matr P_{1}^{2}(\lambda)\right)^2+\left(\matr P_{2}^{2}(\lambda)\right)^2+\left(\matr P_{3}^{2}(\lambda)\right)^2-\left(\matr P_{4}^{3}(\lambda)\right)^2=0
\end{equation*}
This polynomial has at most six roots. 
Only the real parameters should be kept. Thus we discard the complex roots.
The candidate solutions for parameters $\gamma$, $\delta$ and $\epsilon$ are obtained by substituting the estimated real roots for $\lambda$ into Eq.~\ref{eq:x_sol}.

\noindent \textit{Degenerate configurations.} 
The proposed algorithm estimates the two angles using a linear system of equations that consists of a minimum of three independent equations originating from point correspondences. The coefficient matrix can be divided into two parts, the left two columns relate to angle $\alpha$, the third and fourth ones to $\left( \alpha + \beta \right)$. Algebraically, the configuration is degenerate only if one of the parts consists of only zero elements. It is possible only if the 2nd coordinates of all point correspondences are zero.

\section{Experimental results}

In this section, we test the proposed method both on synthesized and publicly available real-world data. The compared methods are the proposed one, the techniques proposed by Choi et al.~\cite{choi2018} and the five-point algorithm of Stewenius et al.~\cite{stewenius2006recent}.

\subsection{Synthetic tests}

\begin{figure*}[h]
  	\centering
  	\includegraphics[width=0.65\columnwidth]{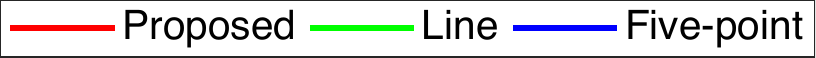}\\[0.3cm]
	\begin{subfigure}[t]{1.328\columnwidth}
    	\centering
  	    \includegraphics[width=0.49\columnwidth]{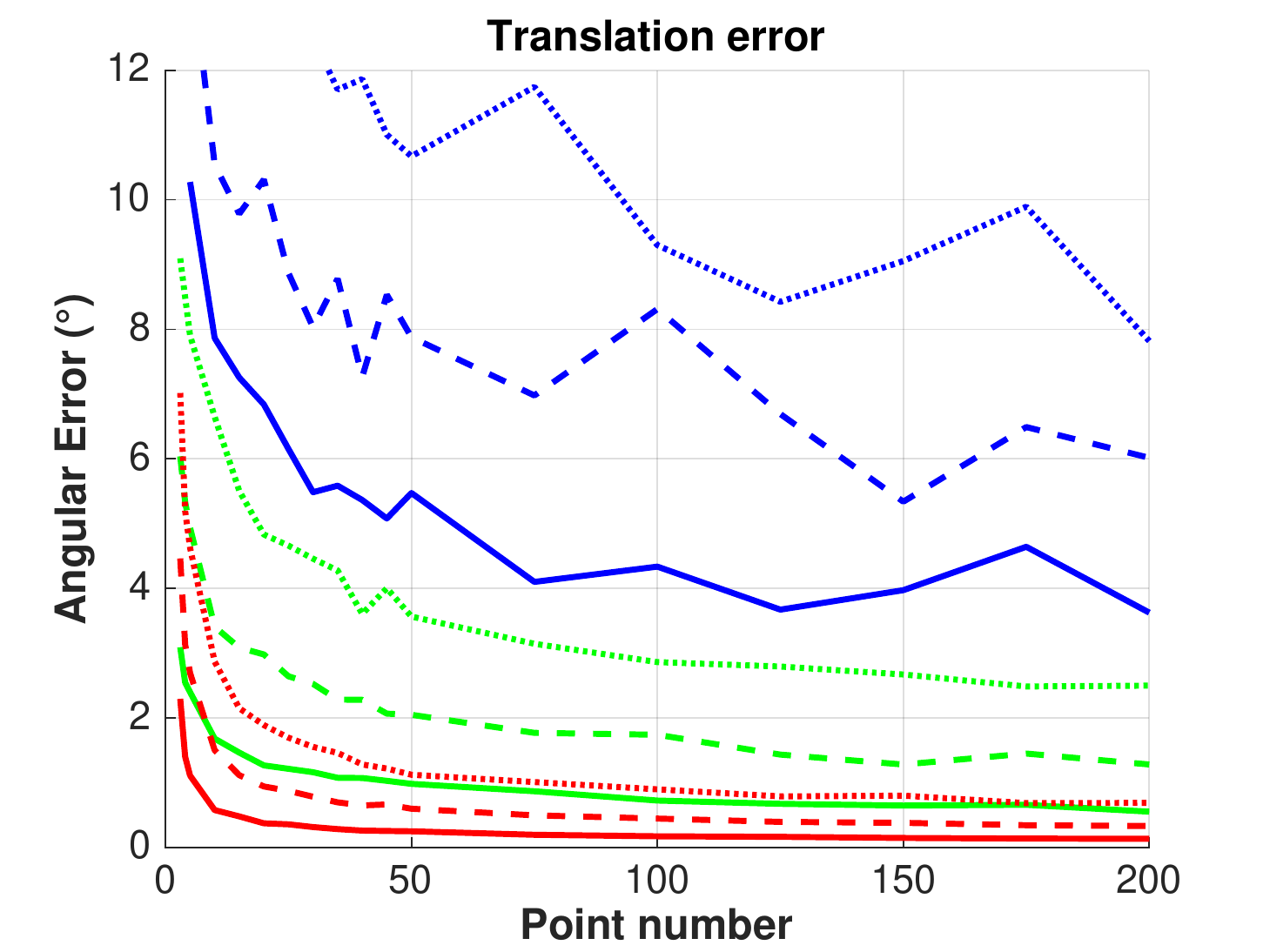}
  	    \includegraphics[width=0.49\columnwidth]{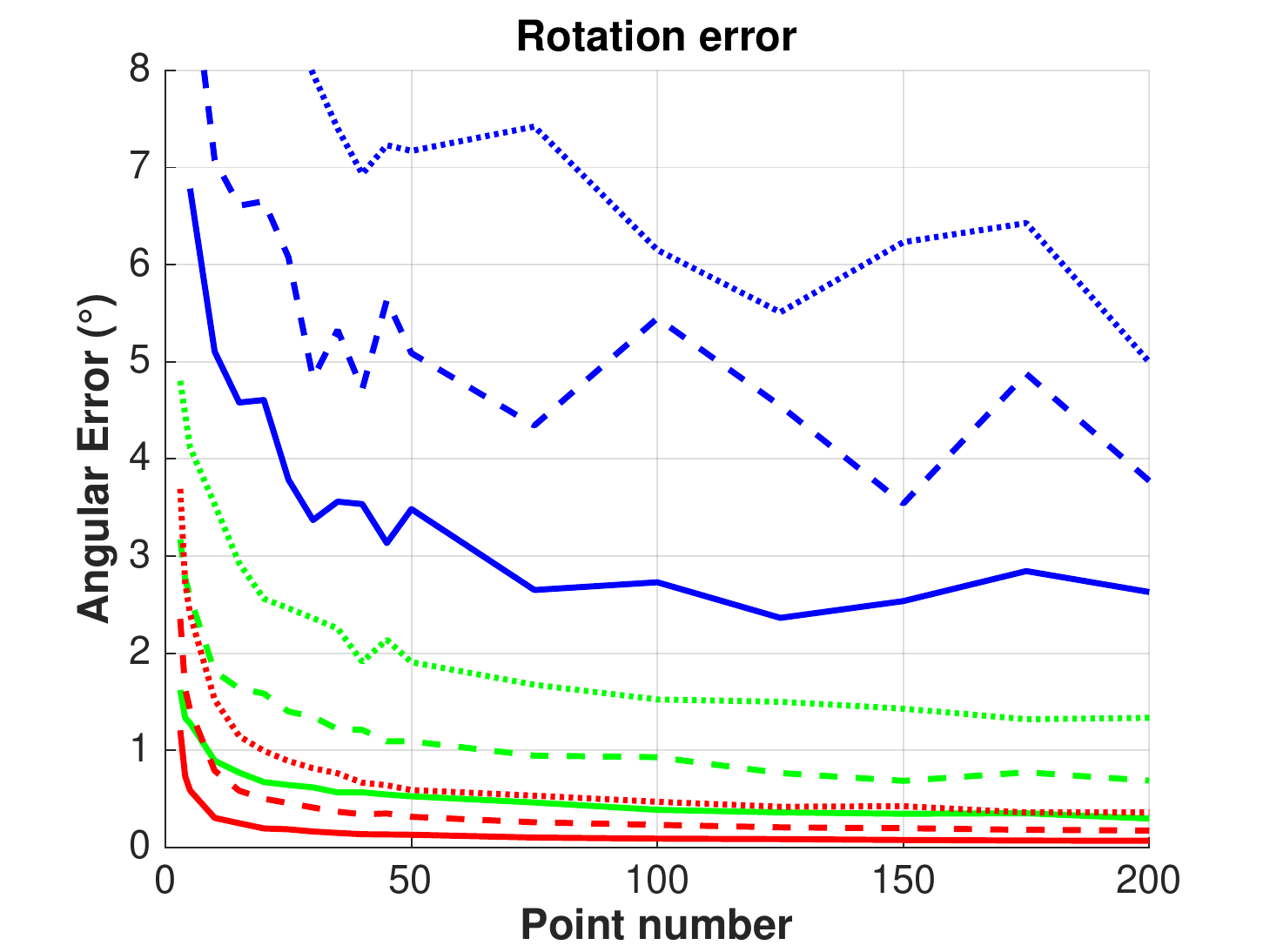}\\[0.2cm]
    	\includegraphics[width=0.75\columnwidth]{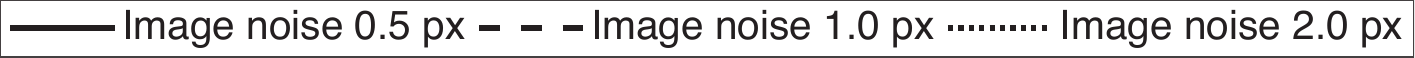}
  	\end{subfigure}
	\begin{subfigure}[t]{0.65\columnwidth}
    	\centering
  	    \includegraphics[width=1.0\columnwidth]{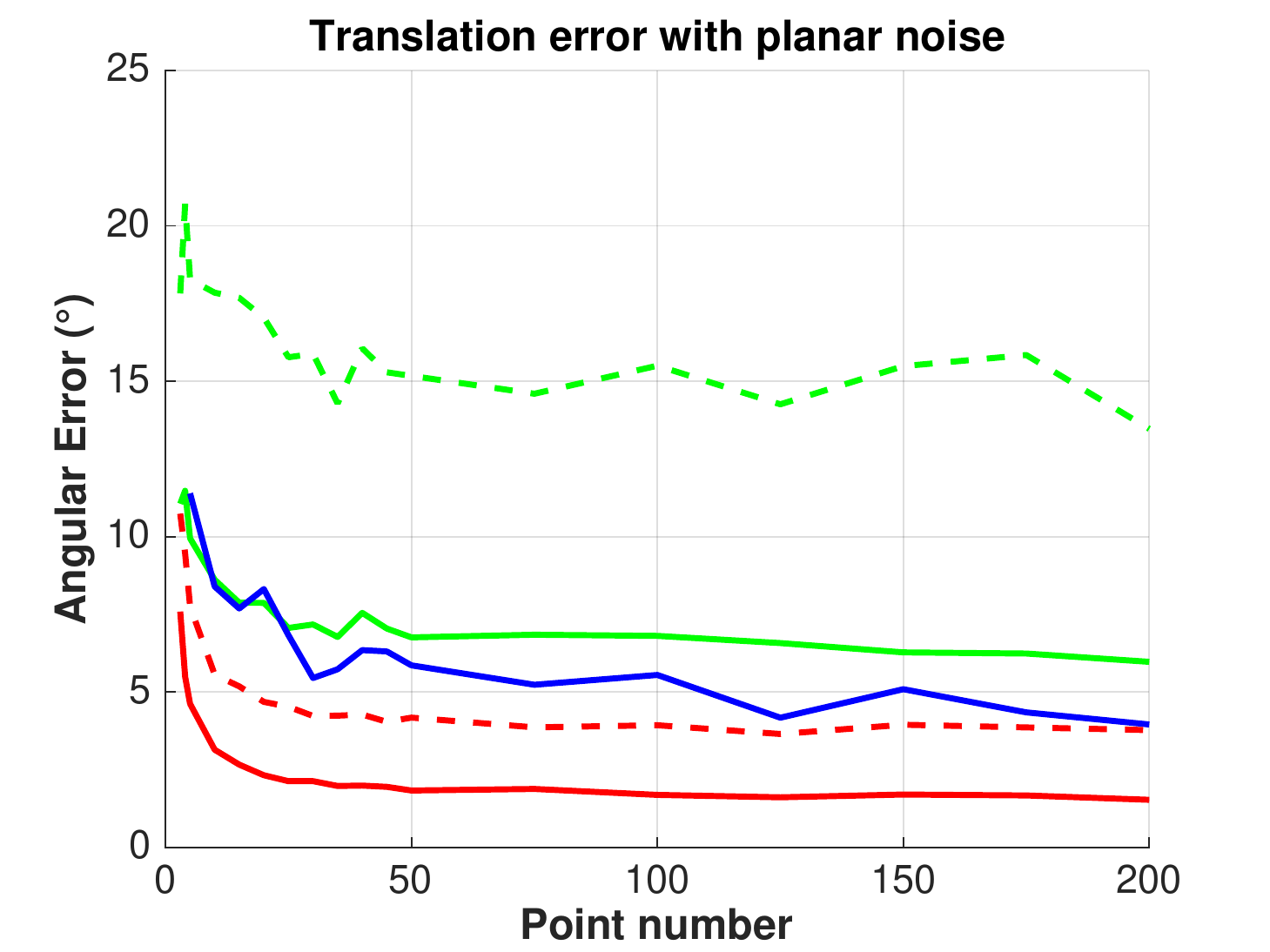}\\[0.2cm]
  	    \includegraphics[width=0.85\columnwidth]{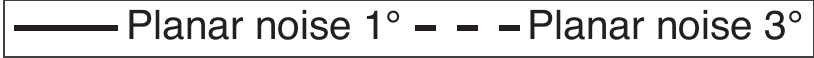}
  	\end{subfigure}
  	\caption{ Relative pose estimation of a purely forward-moving camera in a synthetic environment. In the left two figures, the angular errors (vertical axis; in degrees) of, respectively, the estimated translations (left) and rotations (middle) are plotted as the function of the point number (horizontal axis) used for the estimation with different image noise $\sigma$s (in pixels) added to the point coordinates. 
  	  The right plot reports the translation error (vertical axis; in degrees) as the function of the point number when the camera movement is not entirely planar. In this case, the image noise was set to $0.5$ pixels.
   	  The compared methods are the proposed one, that of Choi et al.~\cite{choi2018} (line) and the five-point algorithm of Stewenius et al.~\cite{stewenius2006recent} (five-point).}
	\label{fig:synth_experiments}
\end{figure*}

To test the proposed method in a fully controlled environment, two cameras were generated by their $3 \times 4$ projection matrices $\textbf{P}_1 = \textbf{K}_1 [\textbf{I}_{3 \times 3} \; | \; 0]$ and $\textbf{P}_2 = \textbf{K}_2 [\textbf{R}_2 \; | -\textbf{R}_2 \textbf{t}_2]$. 
Camera $\textbf{P}_1$ was located in the origin and its image plane was parallel to plane $\text{XY}$. 
The second camera was generated by applying a $10$ unit long purely forward motion to the first one and adding a small random rotation ($\leq 5^\circ$).
Thus $\textbf{t}_2 = 10 \; [\cos(\gamma) \; 0 \; \sin(\gamma)]^\trans$, where $\gamma = 0$. 
Its orientation was determined by a random rotation affecting around axis $\text{Y}$.
Both cameras had a common intrinsic camera matrix with focal length $f_x = f_y = 1000$ and principal points $[500, 500]^\trans$.
3D points were generated at most one unit far from the origin and were projected into both cameras. 
Finally, zero-mean Gaussian-noise with $\sigma$ standard deviation was added to the projected point coordinates. 
For both the obtained rotation and translation, the error is the angular error in degrees. 

The compared methods are the proposed one, that of Choi et al.~\cite{choi2018} and the five point solver of Stewenius et al.~\cite{stewenius2006recent}. The proposed method and Choi et al.~\cite{choi2018} exploit the planar movement when estimating the pose parameters from the over-determined system. The five-point algorithm solves the general pose estimation problem. For applying it to the over-determined problem, the correspondences were fed into the null-space calculation. Finally, the four singular vectors, corresponding to the four smallest singular values, were used as the null-space and the remaining steps of the algorithm were applied using them.

The left two plots of Fig.~\ref{fig:synth_experiments} reports the average -- over $1\;000$ runs on each point number -- rotation (left) and translation (middle) errors plotted as the function of the point number used for the estimation.
The methods were tested on different image noise $\sigma$s (i.e., 0.5, 1.0 and 2.0 pixels; the line style indicates the noise level) added to the point coordinates. 
It can be seen that \textit{the proposed solver is significantly more accurate} than the competitor ones no matter what the image noise level is. This holds for both the estimated rotation and translation.
Also, the algorithm is consistent, i.e., the more points are used, the more accurate the method is. 

For the right plot of Fig.~\ref{fig:synth_experiments}, the planar constraint was corrupted by simulating a small hill with steepness set to $1^\circ$ and $3^\circ$ (the line style indicates the steepness). Therefore, the second camera was rotated around axis X and, also, its Y coordinate was set accordingly. The image noise was fixed to 0.5 pixels. 
It can be seen that the \textit{proposed optimal solver is significantly less sensitive} to the corruption of the planarity constraint than the method of Choi et al.~\cite{choi2018}. With this small planar noise, it leads to more geometrically accurate results than the five-point algorithm. 
Note that, in case of high planar noise, the camera rotation can be corrected by the gravity vector if known. Otherwise, all methods considering planar movement fail.

\begin{figure}[h]
  	\centering
  	\includegraphics[width=0.65\columnwidth]{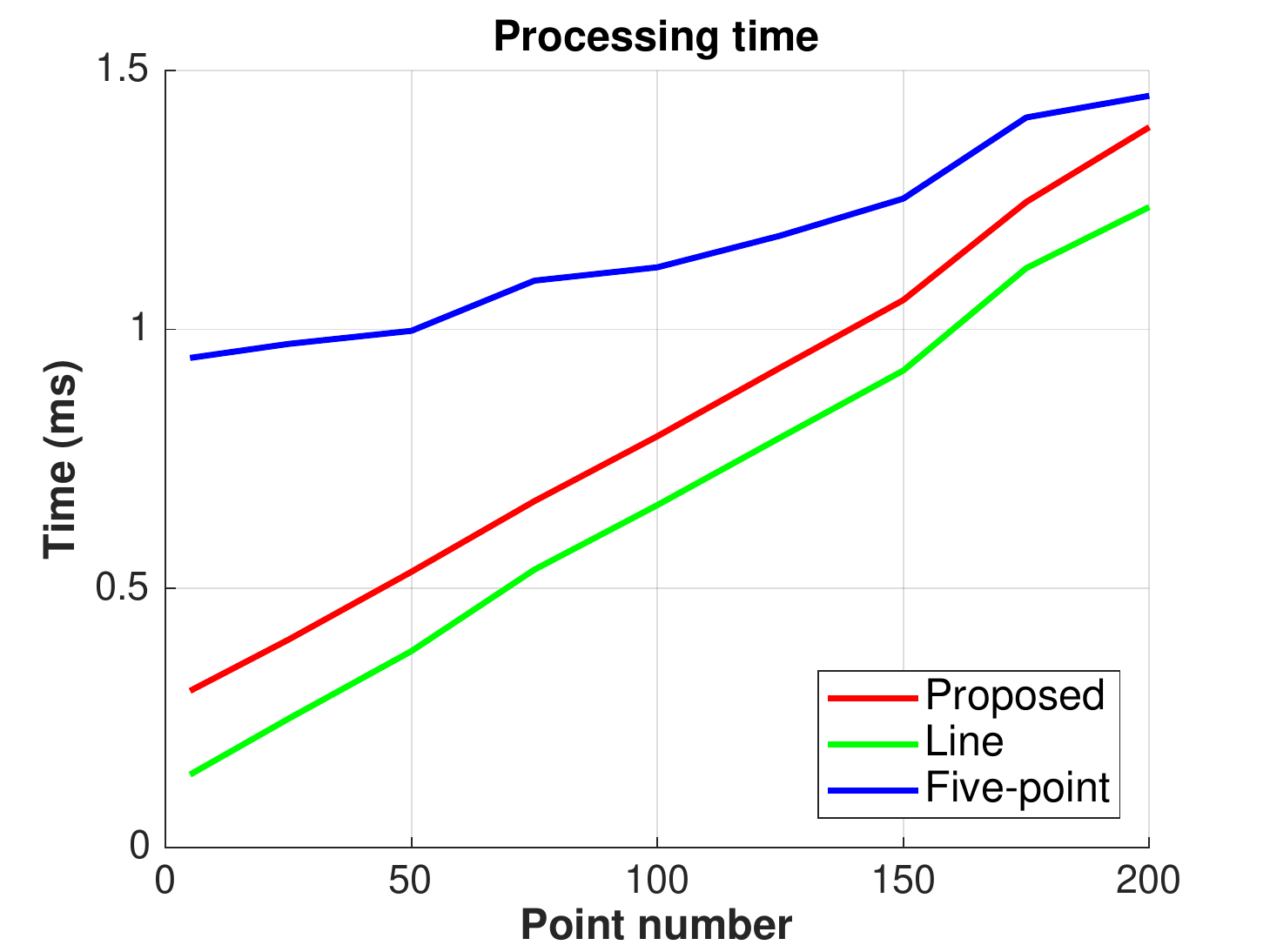}
  	\caption{The processing time (in milliseconds; implemented in Matlab) as the function of the number of points used for the estimation. The compared methods are the proposed one, that of Choi et al.~\cite{choi2018} (line) and the five-point algorithm of Stewenius et al.~\cite{stewenius2006recent} (five-point). }
	\label{fig:processing_times}
\end{figure}

\begin{figure}[h]
  	\centering
  	\includegraphics[width=0.65\columnwidth]{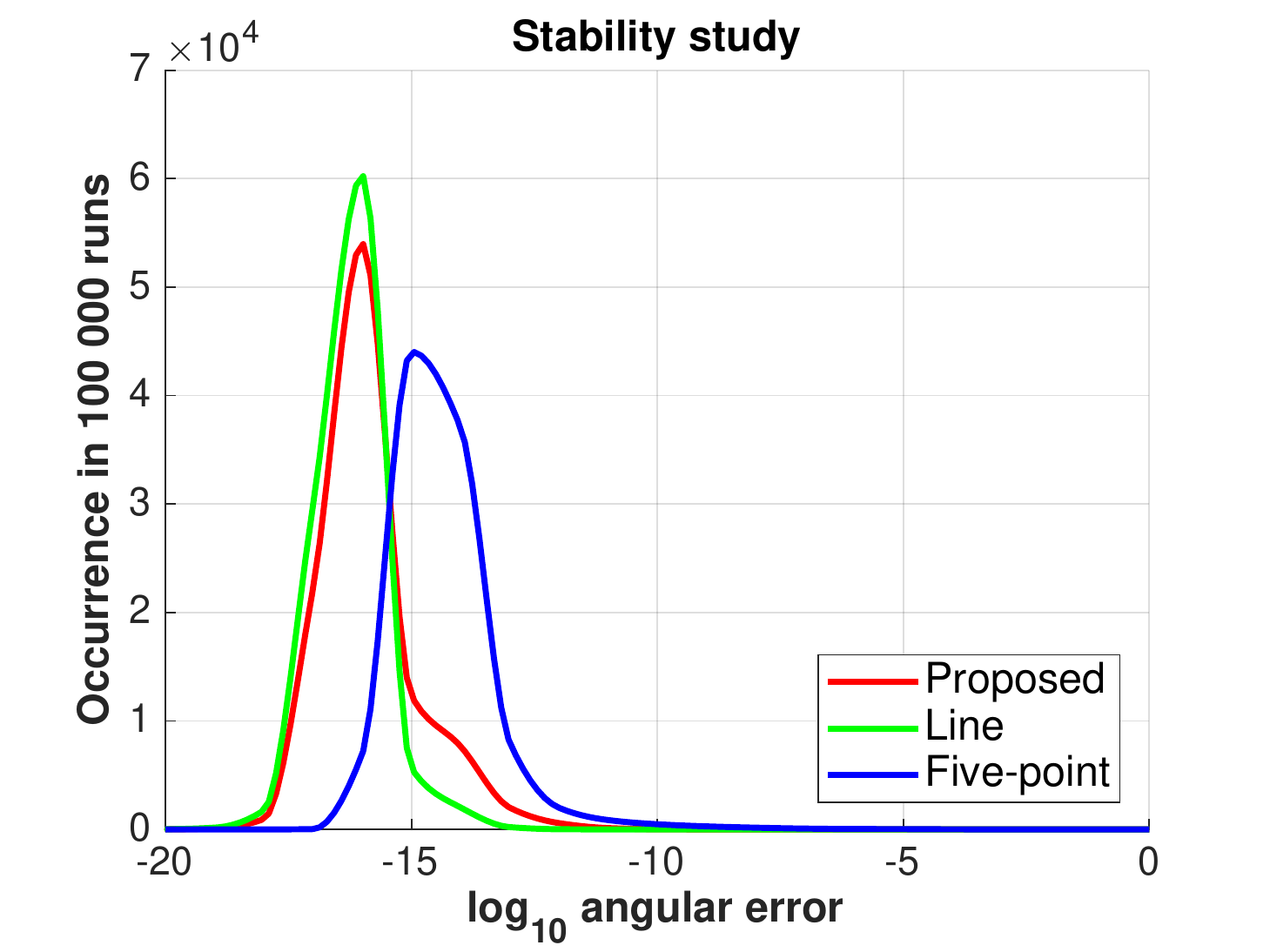}
  	\caption{The number of occurrences (in $100\;000$ runs; vertical axis) of the $\log_{10}$ errors (horizontal) in the estimated rotations in the noise-free case are shown. The compared methods are the proposed one, that of Choi et al.~\cite{choi2018} (line) and the five-point algorithm of Stewenius et al.~\cite{stewenius2006recent}.}
	\label{fig:stability_study}
\end{figure}

In Fig.~\ref{fig:processing_times}, the average (over $100\;000$ runs) processing times (vertical axis; in milliseconds) of the Matlab implementations of the competitor algorithms are reported.
The proposed method is slightly slower than the method of Choi et al.~\cite{choi2018} -- by approx.\ $0.2$ ms --, but is faster than the five-point algorithm by approx.\ $0.5 - 0.6$ ms.

In Fig.~\ref{fig:stability_study}, the number of occurrences (in $100\;000$ runs; vertical axis) of the $\log_{10}$ errors (horizontal) in the estimated rotations (noise-free case) are shown. In each test, the number of points were set randomly from interval $[5, 200]$.
It can be seen that all methods are stable, i.e., there is no peak on the right side of the plot.
The proposed solver and the method of Choi et al.~\cite{choi2018} are slightly more stable than the five-point algorithm.

In summary, the synthetic experiments show that when the movement is close to planar, the proposed solver leads to the most geometrically accurate relative pose estimates in the over-determined case.  
Also, it is of similar speed (slightly slower) as the fastest state-of-the-art alternative, i.e., the method solving the problem as line and ellipse intersection proposed by Choi et al.~\cite{choi2018}. 

\subsection{Real-world experiments}

\begin{figure}[h]
  	\centering
  	\includegraphics[width=0.72\columnwidth]{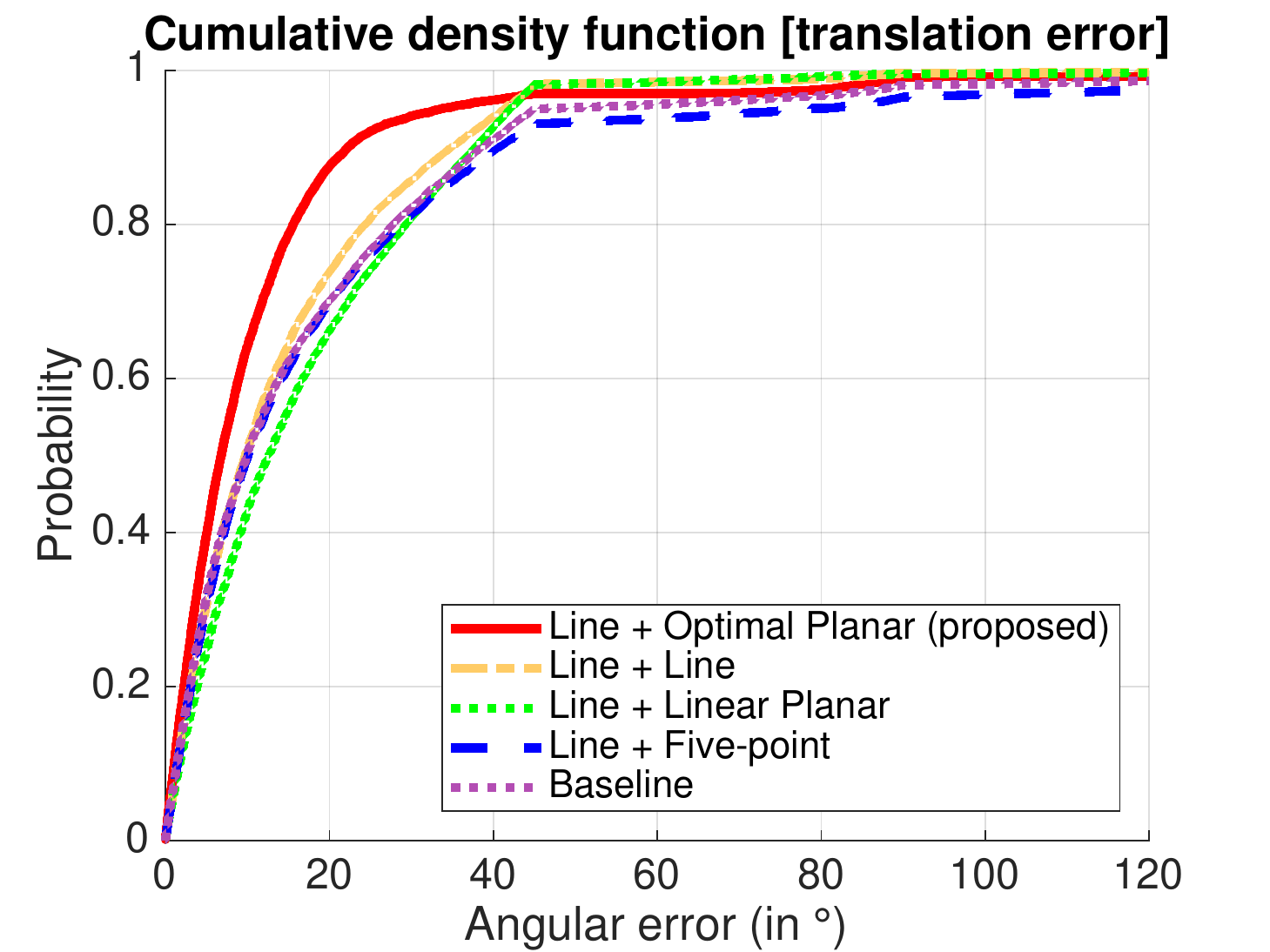}\\[0.2cm] \includegraphics[width=0.72\columnwidth]{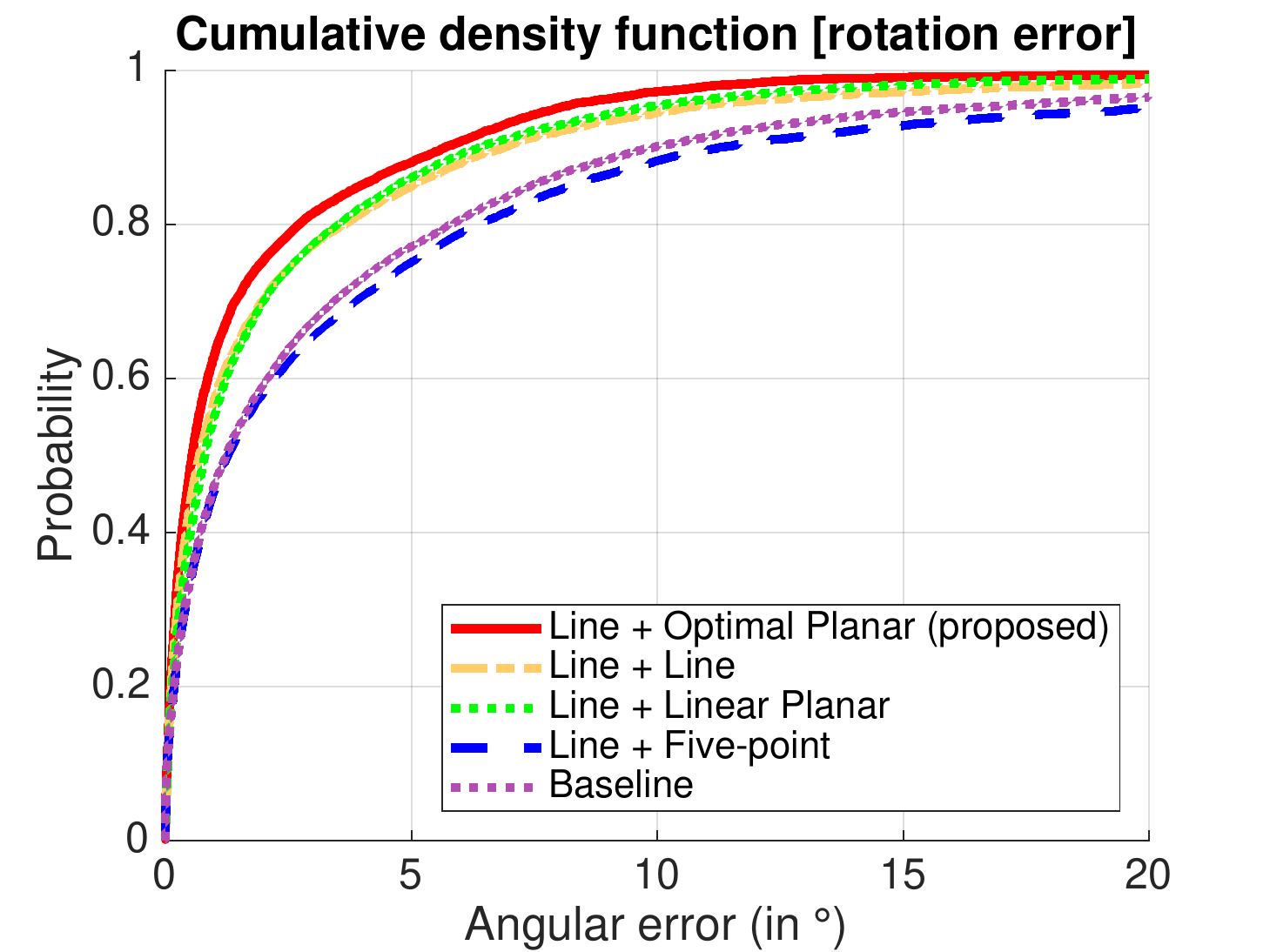}\\[0.2cm] 
  	\includegraphics[width=0.72\columnwidth]{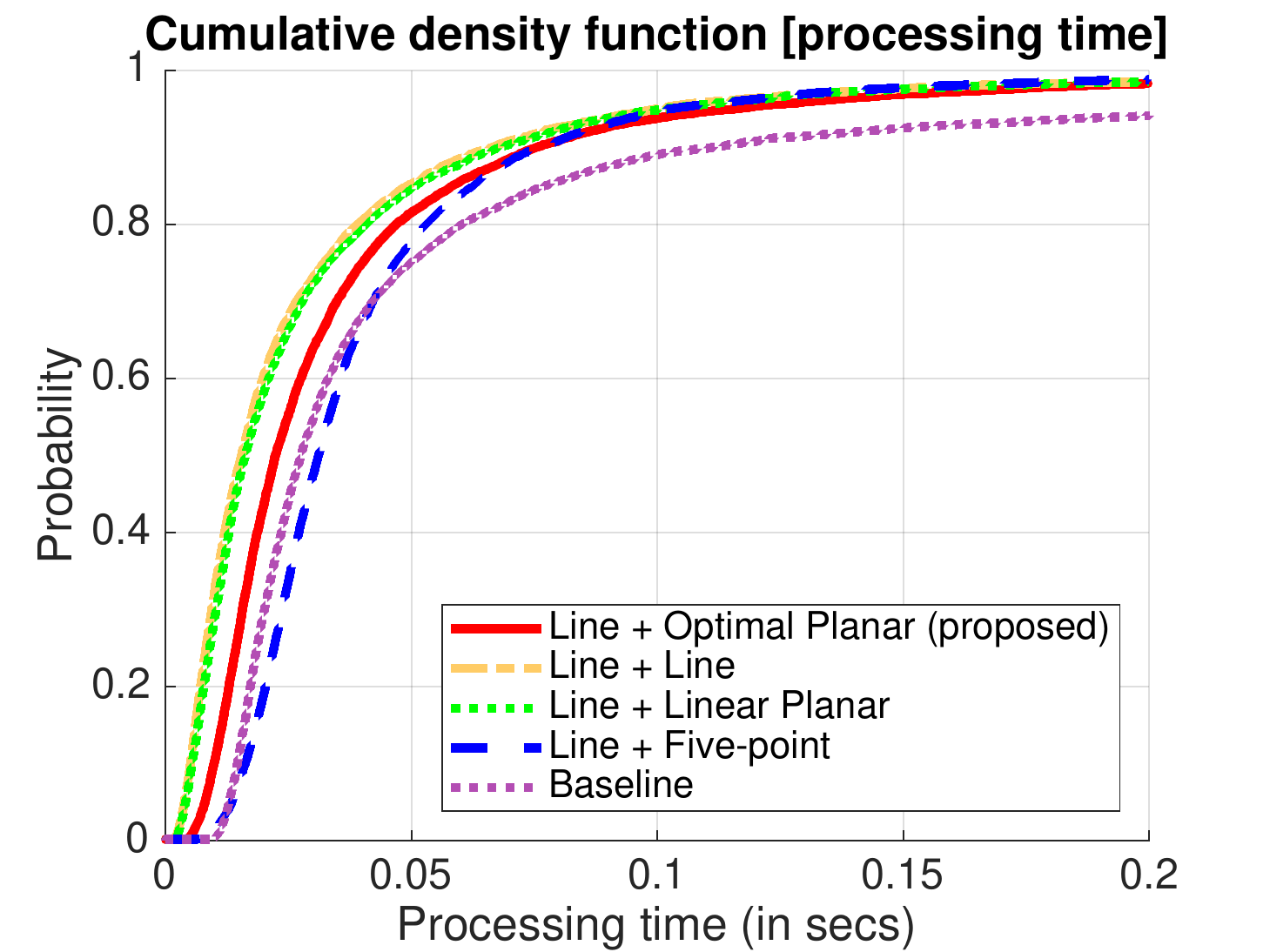}
  	\caption{\textit{Cumulative density functions} of the angular error (top -- translation; middle -- rotation; in degrees) and processing time (bottom; in seconds) on the $15$ scenes ($9\;064$ image pairs in total) of the Malaga dataset. The probabilities (vertical axis) are shown as the function of the angular error and processing time (horizontal). 
  	GC-RANSAC~\cite{barath2017graph} is used as a robust estimator. 
  	A method being accurate is equivalent to its curve being on the left side of the plot.
  	The names of the methods are described in Table~\ref{tab:method_combinations}. }
	\label{fig:real_experiments}
\end{figure}

In order to test the proposed technique on real-world data, we chose the {\fontfamily{cmtt}\selectfont Malaga}\footnote{\url{https://www.mrpt.org/MalagaUrbanDataset}} dataset~\cite{Claraco2014}. 
This dataset was gathered entirely in urban scenarios with a car equipped with several sensors, including one high-resolution stereo camera and five laser scanners. 
We used the sequences of one high-resolution camera and every $10$th frame from each sequence. 
The proposed method was applied to every consecutive image pair. 
The ground truth paths were composed using the GPS coordinates provided in the dataset.
Each consecutive frame-pair was processed independently, therefore, we did not run any optimization minimizing the error on the whole path or detecting loop-closure. The estimated relative poses of the consecutive frames were simply concatenated. The only correction done on the estimated angles ($\in (-180, 180]$) was that we assumed continuous path and, thus, the angles were used with a modulo $90^\circ$. For instance, if the estimated angle was $110^\circ$, we used $20^\circ$, or for $-110^\circ$, it was $-20^\circ$.
In total, $9\;064$ image pairs were used in the evaluation.

%Since the scale, i.e.\ the length of the translation vector, cannot be retrieved due to the ambiguity of perspective projection, we calculated length of the motion from the corresponding GPS coordinates. 
%In total, the solvers were tested on $26.216$ image pairs.

As a robust estimator, we chose Graph-Cut RANSAC~\cite{barath2017graph} (GC-RANSAC) since it is state-of-the-art and its source code is publicly available\footnote{\url{https://github.com/danini/graph-cut-ransac}}.
In GC-RANSAC (and other RANSAC-like methods), two different solvers are used: (a) one for fitting to a minimal sample and (b) one for fitting to a non-minimal sample when doing model polishing on all inliers or in the local optimization step. For (a), the main objective is to solve the problem using as few data points as possible since the processing time is a function of the point number required for the estimation. Except for one case (when we used the five-point algorithm~\cite{stewenius2006recent}), we chose the method called "line" of Choi et al.~\cite{choi2018} which solves the problem from a minimum of two point correspondences and is reported to be extremely fast. 
For (b), we compare the proposed method, the technique solving the planar motion problem as a linear system by the DLT algorithm (discussed in~\cite{choi2018}), solver "line" of Choi et al.~\cite{choi2018}, and the general five-point algorithm~\cite{stewenius2006recent}.
The tested combinations for (a) and (b) are reported in Table~\ref{tab:method_combinations}.

Most of the tested algorithms return multiple pose candidates. 
To select the best one, we did not use $5\%$ of the points (or minimum a single point) in the fitting. 
The pose was estimated from the $95\%$ and, finally, the candidate was selected which minimizes the error on the left out points. 
The only solver returning a single solution and, thus, not requiring this procedure is the planar linear method. 

The accuracy of the compared methods is shown in the top two charts of Fig.~\ref{fig:real_experiments}. The cumulative density functions are reported. 
A method being accurate is equivalent to its curve being on the left side of the plot. 
The top chart shows the accuracy of the estimated translation vectors and the middle one shows that of the rotations.    
For example, for the proposed method (red curve), the probability of returning a translation with lower than $20^\circ$ error is approx.\ $90\%$. 
For all the other solvers, it is around $75\%$.
It can be seen that the \textit{proposed method is significantly more accurate} than the competitor ones. 

The processing time of the whole robust estimation procedure using the compared solvers is shown in the bottom chart of Fig.~\ref{fig:real_experiments}. 
It can be seen that the linear solver (average time is $0.033$ secs) and the method of Choi et al.~\cite{choi2018} (avg.\ is $0.032$ secs) lead to the fastest robust estimation.
GC-RANSAC is marginally slower (avg.\ is $0.039$ secs) when combined with the proposed solver.
However, it is still significantly faster than the estimator using the five-point algorithm -- the average times of "Line + Five-point" and "Baseline" are $0.043$ and $0.065$ seconds, respectively.

In summary, estimating the model parameters from a non-minimal sample by the proposed method leads to accuracy superior to the state-of-the-art without significant overhead in the processing time. 

\begin{table}
\center
  	\resizebox{0.99\columnwidth}{!}{\begin{tabular}{| r || c | c |}
    \hline
    	Name & Minimal solver & Non-minimal solver \\ 
    \hline
 	 	\textbf{Line + Optimal Planar} & planar "line" solver \cite{choi2018} & proposed \\
 	 	\textbf{Line + Line} & planar "line" solver \cite{choi2018} & planar "line" solver \cite{choi2018} \\ % (TODO: cite)
 	 	\textbf{Line + Linear Planar} & planar "line" solver \cite{choi2018} & planar linear solver \cite{choi2018} \\ % (TODO: cite)
 	 	\textbf{Line + Five-point} & planar "line" solver \cite{choi2018} & five-point solver~\cite{stewenius2006recent} \\
 	 	\textbf{Baseline} & five-point solver~\cite{stewenius2006recent} & five-point solver~\cite{stewenius2006recent} \\
    \hline
\end{tabular}}
	\caption{The combinations of solvers used with GC-RANSAC~\cite{barath2017graph} in the real-world experiments (see Fig.~\ref{fig:real_experiments}). The 2nd column shows the solvers used for estimating the pose from a minimal sample. The 3rd column contains the solvers used for polishing the model parameters on a non-minimal sample.}
\label{tab:method_combinations}
\end{table}

\section{Conclusion}

In this paper, a new solver is proposed for estimating the relative camera pose when the camera is moving on a plane, e.g., it is mounted to a car. 
The technique, minimizing an algebraic cost, optimally solves the over-determined case -- i.e., when a larger-than-minimal set of point correspondences is given. 
Therefore, it is extremely useful as the final polishing step of RANSAC~\cite{fischler1981random} or in the local optimization of locally optimized RANSACs~\cite{chum2003locally,barath2017graph}.
The pose parameters are recovered in closed-form as the roots of a $6$th degree polynomial making the procedure fast and stable.  
It is validated both in our synthetic environment and on more than $9,000$ publicly available real image pairs that the method leads results superior to the state-of-the-art in terms of geometric accuracy. 
Even though the solver is marginally slower than the state-of-the-art (by $0.2$ milliseconds on average), it does not lead to noticeably slower robust estimation when combined with GC-RANSAC. 
\textit{The source code of the solver included in GC-RANSAC will be made available after publication.}

\bibliographystyle{unsrt}
\bibliography{egbib}

\end{document}